\documentclass[runningheads]{llncs}
\usepackage[T1]{fontenc}
\usepackage{graphicx,verbatim}
\usepackage{amsmath,amssymb}
\usepackage{hyperref}
\usepackage{siunitx}
\usepackage{xcolor}
\usepackage{algorithm}
\usepackage{booktabs}
\usepackage{algpseudocode}
\usepackage{subcaption}
\usepackage{orcidlink}
\usepackage{bm}
\usepackage{tikz}
\usetikzlibrary {arrows.meta}
\usepackage{multirow}
\algrenewcommand\algorithmicrequire{\textbf{Input:}}
\algrenewcommand\algorithmicensure{\textbf{Output:}}

\usepackage{color}

\begin{document}
\title{AutoFFS: Adversarial Deformations for Facial Feminization Surgery Planning}
\titlerunning{Adversarial Deformations for Facial Feminization Surgery Planning}
%
\author{Paul Friedrich\inst{1}\orcidlink{0000-0003-3653-5624} \and Florentin Bieder\inst{1}\orcidlink{0000-0001-9558-0623} \and Florian M. Thieringer\inst{1,2}\orcidlink{0000-0003-3035-9308} \and Philippe C. Cattin\inst{1}\orcidlink{0000-0001-8785-2713}}
%
\authorrunning{P. Friedrich et al.}
\institute{Department of Biomedical Engineering, University of Basel, Allschwil, Switzerland \and
Department of Oral and Cranio-Maxillofacial Surgery, University Hospital Basel, Basel, Switzerland\\
\email{paul.friedrich@unibas.ch}}
  
\maketitle              
\begin{abstract}
Facial feminization surgery (FFS) is a key component of gender affirmation for transgender and gender diverse patients, aiming to reshape craniofacial structures toward a female morphology. Current surgical planning procedures largely rely on subjective clinical assessment, lacking quantitative and reproducible anatomical guidance. We therefore propose \textbf{AutoFFS}, a novel data-driven framework that generates counterfactual skull morphologies through adversarial free-form deformations. Our method performs a \emph{deformation-based targeted adversarial attack} on an ensemble of pre-trained binary sex classifiers that learned sexual dimorphism, effectively transforming individual skull shapes toward the target sex. The generated counterfactual skull morphologies provide a quantitative foundation for preoperative planning in FFS, driving advances in this largely overlooked patient group. We validate our approach through classifier-based evaluation, propose \emph{Morphological Fr\'{e}chet Distance (MFD)} and \emph{Morphological Kernel Distance (MKD)} to evaluate distributional alignment of generated and real populations, and perform a human perceptual study, confirming that the generated morphologies exhibit target sex characteristics. The project page is available at: \url{https://pfriedri.github.io/autoffs-io}.

\keywords{Facial Feminization Surgery \and Counterfactual Shape Editing \and Adversarial Deformations.}

\end{abstract}
\section{Introduction}
Gender affirmation surgeries enable transgender and gender diverse patients to align their physical appearance with their gender identity.\footnote{Throughout this paper, we distinguish between \textit{sex} and \textit{gender}. Sex refers to biological and anatomical characteristics (such as skeletal morphology), while gender refers to an individual's internal sense of identity and social role.} For many transgender women, facial feminization surgery (FFS) represents a critical component of this process. FFS encompasses a wide range of craniomaxillofacial procedures designed to reshape masculine-appearing facial features into more feminine structures~\cite{tirrell2022facial}. Common procedures include mandibular setback, genioplasty, rhinoplasty, forehead cranioplasty, and fronto-orbital reshaping~\cite{akhavan2021review,becking2007transgender}. These surgeries play an essential role in gender transition and have been shown to significantly improve quality of life for transgender women~\cite{ainsworth2010quality,dang2022evaluation,dubov2018facial}. Despite these documented benefits, surgeons face considerable challenges in preoperative planning. The literature describes a wide variety of surgical techniques, yet selecting appropriate procedures for each patient's unique facial anatomy remains difficult~\cite{kuruoglu2021point,tirrell2022facial}. Current planning approaches largely rely on experience and subjective assessment, leaving room for data-driven tools that could provide more objective guidance. To close this gap, we propose a novel framework for assisting preoperative planning, called \textbf{AutoFFS}, that generates counterfactual skull morphologies, answering the question: \emph{What would this skull look like if it had the opposite sex?} Our approach employs a free-form deformation (FFD)-based \emph{targeted adversarial attack}~\cite{carlini2017towards} on an ensemble of pre-trained binary sex classifiers, effectively deforming a given skull toward the desired sex, and supporting personalized surgical planning grounded in anatomical data. While this study focuses on FFS, which accounts for the majority of facial gender-affirming surgeries, the proposed method could also be applied to facial masculinization surgeries (FMS).
\subsubsection{Related Work} Early work on computer-assisted FFS planning primarily focused on developing virtual surgery planning procedures~\cite{escandon2022applications,rancu2021virtual,sanz2024f,sharaf2022point}, often involving CAD-based methods for preoperative cutting guide design~\cite{ganry2022low,gutierrez2024shaping}. More recent studies have explored automatic planning and cutting guide generation for isolated anatomical regions, such as the mandible~\cite{beyer2025innovations} or the lower jaw~\cite{marin20243d}. Despite these advances, existing methods largely rely on subjective assessment, experience, or reference morphologies~\cite{cronin2024preoperative}, and fail to account for the global morphological interdependencies across the entire skull.

Deep learning based methods have demonstrated remarkable capacity to model discriminative sexual dimorphisms, i.e., different morphological characteristics based on a subject's sex, and have successfully been applied to sex classification tasks \cite{cciftcci2024human,del2023mapping,noel2024sex}. Building upon this foundation, we propose using these learned representations to generate counterfactual skull morphologies.
\section{Method}
\label{sec:method}
\subsubsection{Problem Formulation}
\begin{figure}[htbp]
    \includegraphics[width=\textwidth]{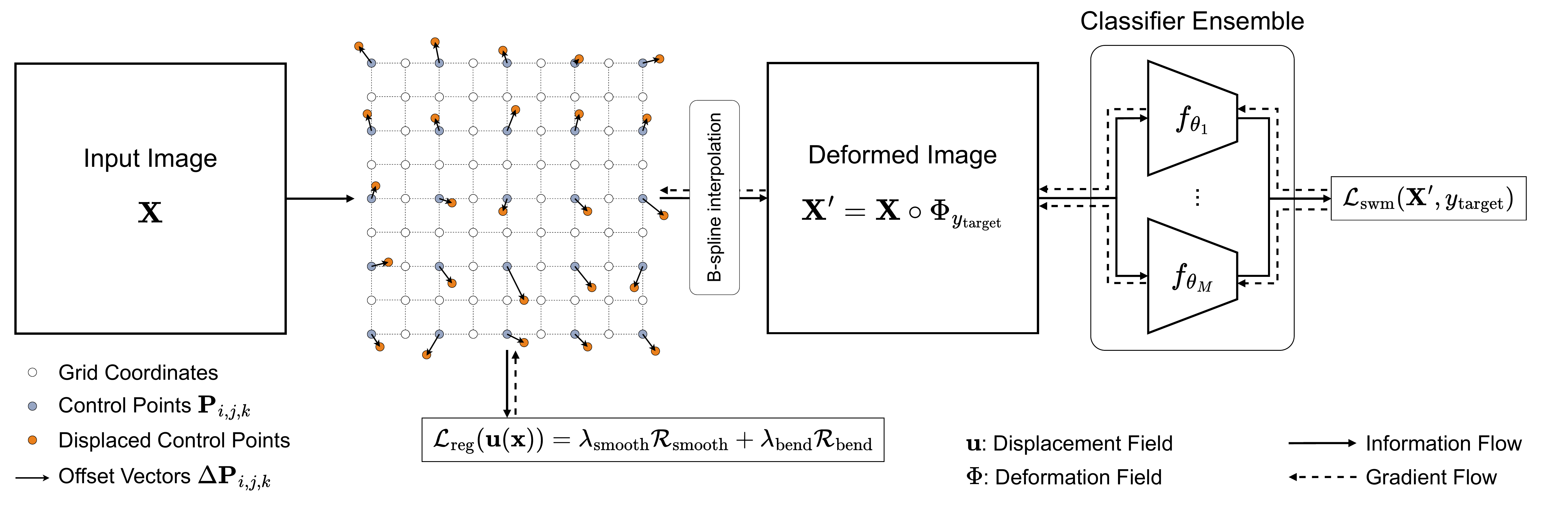}
    \caption{Overview of the proposed pipeline. We perform a \emph{deformation-based targeted adversarial attack} on an ensemble of pre-trained binary sex classifiers by optimizing the deformation field given the target class label $y_{\text{target}}$. Illustration is in 2D for visualization purposes, but the pipeline works in 3D.}
    \label{fig:overview}
\end{figure}
Let $\mathcal{D}$ denote a dataset of skull shapes with $N$ samples $\mathcal{D}:=\left\{(\mathbf{X}_i, y_i)\right\}_{i=1}^{N}$, where each skull is represented by a function $\mathbf{X}_i:\Omega\to\mathbb{R}$ mapping spatial coordinates to occupancy values. Here, $\Omega\subset\mathbb{R}^{3}$ denotes the image domain with discrete dimensions $H\times W\times D$, and $y_i\in\{0,1\}$ denotes a binary sex label: (0) male, (1) female. We propose to solve FFS planning by finding a conditional deformation field $\Phi_{y_{\text{target}}}:\Omega\to\Omega$ that maps each spatial coordinate $\mathbf{x}$ to a new coordinate $\Phi_{y_{\text{target}}}(\mathbf{x})$, such that the transformed skull $\mathbf{X}'$ exhibits features of the target sex $y_{\text{target}}$:
\begin{equation}
\label{eq:apply_def}
    \mathbf{X}' = \mathbf{X} \circ \Phi_{y_{\text{target}}}.
\end{equation}
Here, the composition $(\mathbf{X} \circ \Phi_{y_{\text{target}}})(\mathbf{x}) = \mathbf{X}(\Phi_{y_{\text{target}}}(\mathbf{x}))$ defines the deformed image by resampling the original image at the deformed coordinates. In practice, images are discretized 
on a regular grid, and we use trilinear interpolation to evaluate $\mathbf{X}$ at non-grid positions during the resampling operation. To find such a deformation field $\Phi_{y_{\text{target}}}$, we propose to perform a \emph{targeted adversarial attack}~\cite{carlini2017towards} on an ensemble of pre-trained binary sex classifiers, i.e., we optimize the deformation field such that the transformed scan maximizes the desired class probability. An overview of our proposed approach is shown in \autoref{fig:overview}.

\subsubsection{Classifier Training}
\label{subsec:classification}
To learn a normative representation of the population in our dataset $\mathcal{D}$, we train classification networks to solve a binary sex classification task. Through this task, the networks implicitly learn discriminative features that characterize male and female populations. Formally, each classifier is defined as a neural network $f_\theta: \Omega \mapsto [0,1]$ with parameters $\theta$ that operates on the discretized image representation. The network parameters are trained using Binary Cross-Entropy (BCE) over a batch of $B$ images from the training set
\begin{equation}
    \mathcal{L}_{\text{class}}:= -\frac{1}{B}\sum_{i=1}^{B}\Big[y_i\log(f_\theta(\mathbf{X}_i))+(1-y_i)\log(1-f_\theta(\mathbf{X}_i))\Big].
\end{equation}
To enhance the robustness and representation capacity of our proposed approach, we train an ensemble of $M$ classifiers $\{f_{\theta_m}\}_{m=1}^{M}$ with different architectures. We motivate this choice by the observation that fooling multiple classifiers, each learning distinct feature representations, is substantially more challenging than fooling a single network~\cite{kurakin2018adversarial,pang2019improving}. Consequently, the ensemble approach yields more robust representations that better capture population-level characteristics. Implementation and training details can be found in \autoref{subsec:class_training}.

\subsubsection{Free-Form Deformation}
\label{subsec:ffd}
We parameterize our deformation field $\Phi$ as a 3D cubic B-spline FFD~\cite{rueckert1999nonrigid}. A regular control lattice is defined over the scan domain $\Omega$ with $n_x \times n_y \times n_z$ control points $\mathbf{P}_{i,j,k} \in \mathbb{R}^3$. Each control point has an associated offset $\Delta\mathbf{P}_{i,j,k}$, which defines its displacement from the initial lattice position. The continuous displacement field $\mathbf{u}$ at any spatial location $\mathbf{x}$ is obtained through cubic B-spline interpolation of the control point offsets
\begin{equation}
\label{eq:displacement}
    \mathbf{u}(\mathbf{x}) = \sum_{l=0}^{3}\sum_{m=0}^{3}\sum_{n=0}^{3} B_l(\xi) \, B_m(\eta) \, B_n(\zeta) \, \Delta\mathbf{P}_{i+l,j+m,k+n},
\end{equation}
where $B$ denotes a cubic B-spline basis function, $(i,j,k)$ is the index of the cell containing $\mathbf{x}$, and $(\xi, \eta, \zeta) \in [0,1)^3$ are the local coordinates within that cell. The resulting deformation field is then defined as 
\begin{equation}
\label{eq:deformation}
    \Phi(\mathbf{x}) = \mathbf{x} + \mathbf{u}(\mathbf{x}).
\end{equation}
Given any differentiable loss $\mathcal{L}$ that depends on the deformed image $\mathbf{X}' = \mathbf{X} \circ \Phi$, the gradient with respect to a control-point offset follows from the chain rule
\begin{equation}
\label{eq:ffd_grad}
    \frac{\partial \mathcal{L}}{\partial \Delta\mathbf{P}_{i,j,k}}
    = \sum_{\mathbf{x}\in\Omega}
    B_l(\xi) \, B_m(\eta) \, B_n(\zeta) \;
     \frac{\partial \mathcal{L}}{\partial \mathbf{X}'(\mathbf{x})}\;
    \nabla \mathbf{X}\!\left(\Phi(\mathbf{x})\right),
\end{equation}
where $(l,m,n)$ are the basis function indices corresponding to control point $(i,j,k)$ within the local $4\times4\times4$ support region of $\mathbf{x}$, $\frac{\partial \mathcal{L}}{\partial \mathbf{X}'(\mathbf{x})}$ is the gradient propagated from the loss to the deformed image intensity at location $\mathbf{x}$, and $\nabla \mathbf{X}(\Phi(\mathbf{x}))$ is the spatial gradient of the original image evaluated at the warped coordinate. This enables end-to-end differentiation of the loss $\mathcal{L}$ with respect to all control-point offsets. We will make use of this property when optimizing our deformation field at test time.

\subsubsection{Regularizing the Deformation Field}
\label{subsec:regularization}
While adversarial optimization can effectively drive the deformation process, it is likely to produce unrealistic, non-smooth transformations if unconstrained. We therefore apply two regularization terms: (1) A \emph{smoothness regularizer}, proposed by Balakrishnan et al.~\cite{balakrishnan2019voxelmorph}, that penalizes the Jacobian of the displacement field
\begin{equation}
\label{eq:smoothness_reg}
    \mathcal{R}_{\text{smooth}} = \frac{1}{|\Omega|}
    \sum_{\mathbf{x}\in\Omega} 
    \left\|\frac{\partial \mathbf{u}(\mathbf{x})}{\partial \mathbf{x}}\right\|_F^2,
\end{equation}
where $\|\cdot\|_F$ denotes the Frobenius norm, and (2) a \emph{bending energy regularizer} proposed by Rueckert et al.~\cite{rueckert1999nonrigid} that can be formulated as the discrete, averaged squared Frobenius norm of the Hessian of $\mathbf{u}(\mathbf{x})$
\begin{equation}
\label{eq:bending_reg}
    \mathcal{R}_{\text{bend}} = \frac{1}{|\Omega|}
    \sum_{\mathbf{x}\in\Omega}\left(
    \sum_{p\in\{x,y,z\}}
    \left\|
    \frac{\partial^2 \mathbf{u}(\mathbf{x})}{\partial p^2}
    \right\|_2^2
    +
    2 \sum_{\substack{p,q\in\{x,y,z\}\\ p<q}}
    \left\|
    \frac{\partial^2 \mathbf{u}(\mathbf{x})}{\partial p\,\partial q}
    \right\|_2^2\right).
\end{equation}
Both regularization terms penalize rapid spatial variations in the displacement field, encouraging smooth, physically plausible transformations. We approximate both terms using finite differences on the deformation grid.

\subsubsection{Deformation-Based Targeted Adversarial Attack}
\label{subsec:attack}
\begin{figure}[htbp]
    \includegraphics[width=\textwidth]{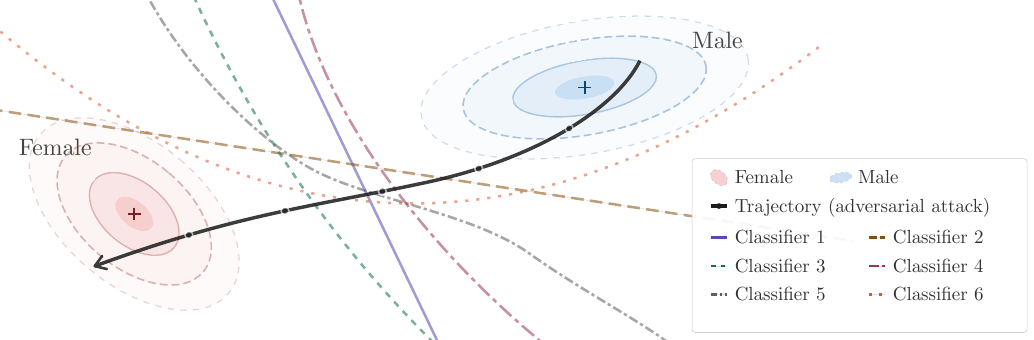}
    \caption{Illustration of the targeted adversarial attack as a trajectory through a simplified 2D representation space. The deformation field optimization moves a sample from the male distribution across all ensemble classifier decision boundaries into the female distribution.}
    \label{fig:attack}
\end{figure}

To apply our method at test time, we perform a \emph{targeted adversarial attack} on the pre-trained classifier ensemble $\{f_{\theta_m}\}_{m=1}^M$ by optimizing our deformation field $\Phi_{y_{\text{target}}}$ to produce a deformed scan $\mathbf{X}'$ that fools these classifiers into predicting the target class label $y_\text{target}$. We therefore freeze all classifier weights and only optimize the control point offsets $\Delta\mathbf{P}_{i,j,k}$, which are initialized to zero (i.e., an identity transform). To optimize these control point offsets, we introduce a \emph{smooth worst-case margin loss} $\mathcal{L}_{\text{swm}}$ based on the log-sum-exp operator \cite{boyd2004convex}, which smoothly approximates the worst-case logit across the ensemble, targeting the most resistant classifier while maintaining gradient flow through all ensemble members. The optimization jointly drives all classifiers beyond a specified logit-margin $\gamma$ through a single unified objective. The temperature parameter $\tau$ controls the smoothness of the approximation, where lower values approach the true min/max. To enforce symmetric deformations, we additionally apply the classifier ensemble to deformed scans mirrored across the midsagittal plane (denoted with $\texttt{flip}(\cdot)$). The whole process is described in \autoref{alg:deformation}.
\begin{algorithm}[htbp!]
\caption{Targeted Adversarial Attack at Test Time}
\label{alg:deformation}
\begin{algorithmic}[1]
\Require Input image $\mathbf{X}$, target class $y_{\text{target}}$, classifier ensemble $\{f_{\theta_m}\}_{m=1}^M$, iterations $S$, logit-margin $\gamma$, temperature $\tau$, learning rate $\alpha$, regularization weights $\lambda_{\text{smooth}}, \lambda_{\text{bend}}$
\Ensure Deformed image $\mathbf{X}'$
\State $\Delta\mathbf{P}_{i,j,k} \leftarrow \mathbf{0}$ \Comment{Initialize control point offsets}
\For{$s = 1$ to $S$}
    \State $\mathbf{u} \leftarrow \texttt{BSplineInterpolate}(\Delta\mathbf{P})$ \Comment{Compute displacement field (Eq. \eqref{eq:displacement})}
    \State $\Phi_{y_{\text{target}}} \leftarrow \mathbf{x} + \mathbf{u}(\mathbf{x})$ \Comment{Compute deformation field (Eq. \eqref{eq:deformation})}
    \State $\mathbf{X}' \leftarrow \mathbf{X} \circ \Phi_{y_{\text{target}}}$ \Comment{Apply deformation (Eq. \eqref{eq:apply_def})}
    \State $\{l_m\}_{m=1}^{2M} \leftarrow \{f_{\theta_m}(\mathbf{X}'), f_{\theta_m}(\texttt{flip}(\mathbf{X}'))\}_{m=1}^M$ \Comment{Obtain ensemble logits}
    \If{$y_{\text{target}} = 1$} \Comment{FFS - Feminization}
        \State $\tilde{l}_{\min} \leftarrow -\tau \log \sum_{m=1}^{2M} \exp(-l_m / \tau)$ \Comment{Smooth minimum}
        \State $\mathcal{L}_{\text{swm}} \leftarrow \max(0, \gamma - \tilde{l}_{\min})$
    \Else \Comment{FMS - Masculinization}
        \State $\tilde{l}_{\max} \leftarrow \tau \log \sum_{m=1}^{2M} \exp(l_m / \tau)$ \Comment{Smooth maximum}
        \State $\mathcal{L}_{\text{swm}} \leftarrow \max(0, \gamma + \tilde{l}_{\max})$
    \EndIf
    \State $\mathcal{L}_{\text{total}} \leftarrow \mathcal{L}_{\text{swm}} + \lambda_{\text{smooth}} \mathcal{R}_{\text{smooth}} + \lambda_{\text{bend}} \mathcal{R}_{\text{bend}}$ \Comment{Aggregate total loss}
    \State Backpropagate $\mathcal{L}_{\text{total}}$ to obtain $\nabla_{\Delta\mathbf{P}} \mathcal{L}_{\text{total}}$ \Comment{Eq. \eqref{eq:ffd_grad}}
    \State $\Delta\mathbf{P} \leftarrow \texttt{Adam}(\Delta\mathbf{P}, \nabla_{\Delta\mathbf{P}} \mathcal{L}_{\text{total}}, \alpha)$ \Comment{Update control point offsets}
\EndFor
\State \Return $\mathbf{X}'$
\end{algorithmic}
\end{algorithm}
Intuitively, the targeted adversarial attack can be viewed as a tool for traversing a learned representation space. Let's consider a skull that initially lies within the male distribution. By optimizing the deformation field, we progressively move this skull across the decision boundaries of all ensemble classifiers until it is confidently placed within the female distribution (and vice versa for masculinization). The logit-margin $\gamma$ ensures the final sample lies well beyond all boundaries rather than at their edge. A simplified 2D example of this process is visualized in \autoref{fig:attack}.
As the deformation is parameterized by spatially localized control points, displacements can be masked region-wise, allowing clinically or patient-motivated constraints to be imposed directly on the deformation field.
\section{Experiments}
\subsubsection{Dataset}
\label{subsect:dataset}
While most FFS planning approaches rely on computed tomography (CT) imaging, ethical and practical constraints limit access to full-skull CT scans from healthy individuals. We therefore utilize magnetic resonance (MR) scans of multiple sclerosis (MS) patients from the SMSC study~\cite{disanto2016swiss}. As MS is a demyelinating disease of the central nervous system, its pathology is confined to the brain and spinal cord, leaving cranial bone morphology unaffected. This makes the dataset a suitable surrogate for studying skull shape variation. We rely on T1-weighted scans as they are the standard structural sequence acquired across clinical and research protocols, and recent work has shown that cranial surfaces can be recovered from routine T1 acquisitions with sub-millimeter agreement to CT-derived models~\cite{ten2025towards}.

The dataset comprises $444$ MR scans ($152$ male, $292$ female). Since our analysis targets skull morphology, we preprocess all scans by first segmenting bone structures using a pre-trained model from GRACE~\cite{stolte2024precise}. Each scan is then reoriented to RAS+ convention, padded to $256 \times 256 \times 256$ voxels, and cropped to the frontal half ($256 \times 256 \times 128$) to isolate facial features. This focus on the anterior skull is motivated both clinically and perceptually: facial feminization surgery targets anterior structures \cite{dang2022evaluation}, and sex perception in social interaction is predominantly driven by facial features \cite{garvin2012sexual,gonzalez2022sex}. The dataset is partitioned into training ($70\%$), validation ($15\%$), and test ($15\%$) sets.
\subsubsection{Classifier Training \& Performance}
\label{subsec:class_training}
Our proposed framework relies on accurate sex classification to guide the optimization process. We therefore trained a diverse set of models, including four ResNets~\cite{he2016deep} (\texttt{ResNet\{18,34,50,101\}}) and four Squeeze-and-Excitation (SE)-ResNets~\cite{hu2018squeeze} (\texttt{SE-ResNet\{18,34,50,101\}}). All networks were trained with a batch size of 16 for 100 epochs using AdamW~\cite{loshchilov2017decoupled} with a learning rate of $10^{-4}$. To improve generalization, we applied data augmentation during training, including horizontal flipping ($p=0.5$) and affine transformations ($p=1.0$) with scaling $s \sim \mathcal{U}(0.9, 1.1)$, rotations $r_x, r_y, r_z \sim \mathcal{U}(-5^{\circ}, 5^{\circ})$, and translations $t_x, t_y, t_z \sim \mathcal{U}(-10, 10)$ voxels. As we require localized features to effectively drive the optimization process, we apply a \emph{masking strategy} during training. We split the volume into $64 \times 64 \times 32$ sized patches and randomly mask \SI{50}{\percent} of them in each iteration. To account for class imbalance, we weight the BCE loss by the inverse class frequencies. All models were trained on the same dataset split, with model selection based on validation accuracy.
\begin{table}[htbp]
    \centering
    \caption{Classifier performance on the hold-out test set.}
    \begin{tabular}{l|ccc|ccc}
        \toprule
        & \multicolumn{3}{c|}{\texttt{ResNet}} & \multicolumn{3}{c}{\texttt{SE-ResNet}} \\
        \textbf{Depth} & \textbf{Acc.} $(\uparrow)$ & \textbf{F1} $(\uparrow)$ & \textbf{AUROC} $(\uparrow)$ & \textbf{Acc.} $(\uparrow)$ & \textbf{F1} $(\uparrow)$ & \textbf{AUROC} $(\uparrow)$ \\\midrule
        \texttt{18}  & $0.924$ & $0.932$ & $0.969$ & $0.894$ & $0.904$ & $0.964$ \\
        \texttt{34}  & $0.939$ & $0.946$ & $0.979$ & $0.924$ & $0.932$ & $0.988$ \\
        \texttt{50}  & $0.924$ & $0.929$ & $0.976$ & $0.848$ & $0.865$ & $0.949$ \\
        \texttt{101} & $0.864$ & $0.880$ & $0.931$ & $0.818$ & $0.842$ & $0.937$ \\\bottomrule
    \end{tabular}
    \label{tab:pclassscores}
\end{table}
We first evaluate the trained classifiers performance on a hold-out test set. We measure accuracy, F1 and AUROC scores and report them in \autoref{tab:pclassscores}. Overall, we find strong classification performance across architectures, with most models achieving accuracies above $0.85$ and AUROC scores consistently exceeding $0.93$. These scores are in line with recently published work~\cite{cciftcci2024human,noel2024sex} and support the claim that the representations learned by our classification networks capture sexual dimorphic features.

\subsection{System-Level Performance}
\label{system}
Evaluating our method poses an inherent challenge, as no paired data exists that would permit a direct comparison between generated and ground-truth morphologies. One alternative would be to compare the generated morphologies to actual post-operative outcomes. However, this conflicts with a core premise of our work, which is not designed to replicate current surgical practice. Instead, following evaluation paradigms from counterfactual generation, we propose four complementary assessments: (1) we investigate whether generated morphologies fool hold-out classification networks, introducing a learned per-sample metric; (2) we propose \emph{Morphological Fr\'{e}chet Distance} (MFD) and \emph{Morphological Kernel Distance} (MKD) that measure distributional alignment between the generated and real target populations; (3) we examine whether the induced morphological changes are consistent with reported sexual dimorphisms in the literature; and (4) we conduct a human perceptual study to investigate whether the morphological changes also fool human observers into perceiving the intended sex.

\subsubsection{Implementation Details}
We divide the classifiers into two groups: (1) classifiers used during optimization (\texttt{ResNet\{18,34,50\}}, \texttt{SE-ResNet\{34,50,101\}}) and (2) hold-out classifiers reserved for evaluation (\texttt{ResNet101}, \texttt{SE-ResNet18}). The evaluation classifiers were deliberately chosen to differ in architecture from those used during optimization, ensuring that our approach generalizes beyond the specific learned representations it was trained against. We further define a control point lattice with $n_x, n_y, n_z = 32$ and optimize for $S=100$ steps using Adam~\cite{kingma2014adam} with a learning rate of $\alpha=5\times10^{-3}$ and a cosine annealing schedule~\cite{loshchilov2016sgdr}. We set $\lambda_\text{smooth}, \lambda_\text{bend} = 1\times10^{8}$, $\gamma=4.5$, and $\tau=1$.\footnote{In the absence of ground-truth target morphologies, hyperparameters were selected based on qualitative assessment of anatomical plausibility.} To ensure modifications are limited to facial features, we fix control point displacements in the posterior half of the volume to zero.

\subsubsection{Classifier-Based Evaluation}
\begin{figure}[htbp]
    \centering
    \begin{tikzpicture}
        \node[anchor=south west] (image) at (0,0) {\includegraphics[width=.9\linewidth]{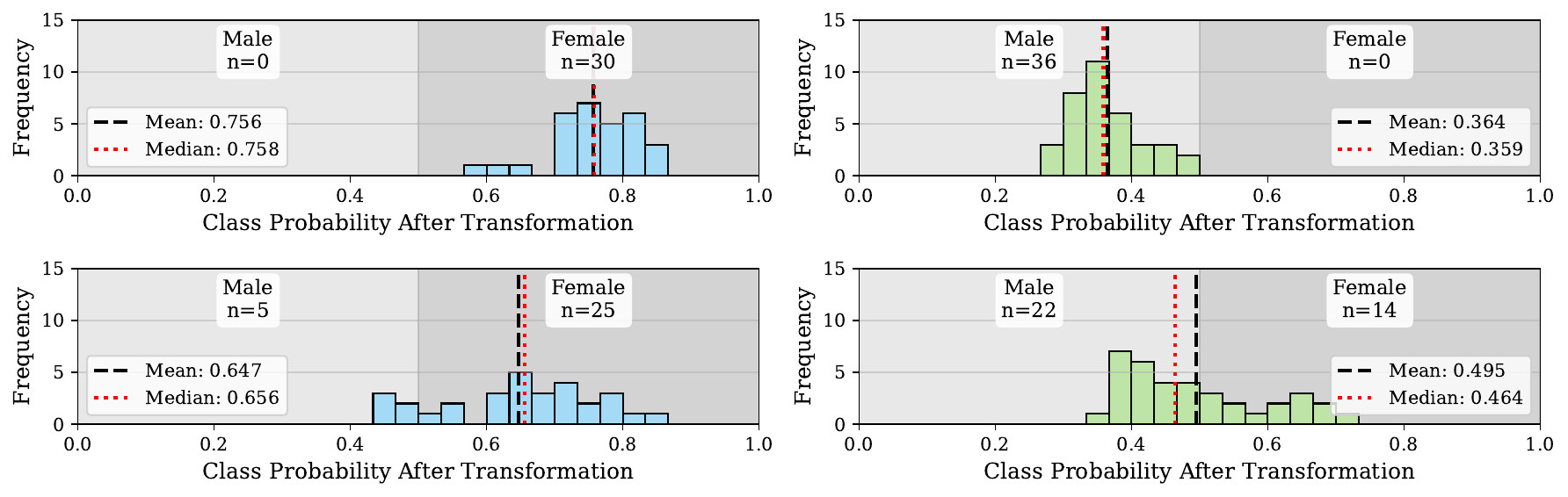}};
        \begin{scope}[x={(image.south east)}, y={(image.north west)}]
            \node[anchor=west, font=\tiny, align=center] at (0.975, 0.73) {With\\ensembling};
            \node[anchor=west, font=\tiny, align=center] at (0.975, 0.31) {Without\\ensembling};
        \end{scope}
    \end{tikzpicture}
    \caption{Class probabilities from the hold-out classifier ensemble after applying our proposed framework with (\textit{top}) and without (\textit{bottom}) the ensembling strategy.}
    \label{fig:classprobs}
\end{figure}

For the classifier-based evaluation, we first transform the entire test set and evaluate the resulting class probabilities using the hold-out classifiers. As shown in \autoref{fig:classprobs} (\textit{top}), our approach consistently produces deformed scans that are classified with the target sex label, achieving a flip rate of $\text{100}\%$. We additionally report results when only using the single best classifier (\texttt{ResNet34}) without ensembling during optimization (\textit{bottom}), which only results in a flip rate of $\sim71\%$. This comparison highlights the improved consistency afforded by the ensembling strategy, which reduces the variance of the resulting class probabilities, yields more robust deformations and an overall higher flip rate.

\subsubsection{Distributional Alignment}
To complement the per-sample classifier evaluation, we introduce two population-level metrics that quantify how well the distribution of generated morphologies aligns with the target sex distribution. Both operate on penultimate-layer feature embeddings extracted from a hold-out classification network, but make different distributional assumptions and therefore serve as mutual robustness checks.

The first, which we term \emph{Morphological Fr\'{e}chet Distance} (MFD), is analogous to the Fr\'{e}chet Inception Distance (FID)~\cite{heusel2017gans} used in image generation, and is defined as the Fr\'{e}chet distance between two multivariate Gaussians fitted in the learned feature space
\begin{equation}
    \text{MFD} = \|\mathbf{\mu}_g - \mathbf{\mu}_t\|_2^2 + 
    \operatorname{Tr}\!\left(\mathbf{\Sigma}_g + \mathbf{\Sigma}_t - 
    2\left(\mathbf{\Sigma}_g \mathbf{\Sigma}_t\right)^{1/2}\right),
\end{equation}
\noindent where $(\mathbf{\mu}_g, \mathbf{\Sigma}_g)$ and $(\mathbf{\mu}_t, \mathbf{\Sigma}_t)$ denote the mean and covariance of the feature embeddings for the generated and real target populations, and $\operatorname{Tr}(\cdot)$ denotes the matrix trace.

The second, which we term \emph{Morphological Kernel Distance} (MKD), is analogous to the Kernel Inception Distance 
(KID)~\cite{binkowski2018demystifying} and estimates the squared Maximum Mean Discrepancy (MMD) between the two populations using an unbiased U-statistic with a polynomial kernel $k(\mathbf{x}, \mathbf{y}) = \left(\tfrac{1}{d}\,\mathbf{x}^\top \mathbf{y} + 1\right)^3$ \cite{gretton2012kernel}
\begin{equation}
    \text{MKD} = 
    \frac{1}{n(n-1)}\sum_{i\neq j}^{n} 
    k(\mathbf{x}_i, \mathbf{x}_j)
    + \frac{1}{m(m-1)}\sum_{i\neq j}^{m} 
    k(\mathbf{y}_i, \mathbf{y}_j)
    - \frac{2}{n m}\sum_{i=1}^{n}\sum_{j=1}^{m} 
    k(\mathbf{x}_i, \mathbf{y}_j),
\end{equation}
\noindent where $\{\mathbf{x}_i\}_{i=1}^{n}$ and $\{\mathbf{y}_j\}_{j=1}^{m}$ denote the feature embeddings of the generated and real target populations (with $n$ and $m$ denoting their respective sample sizes), and $d$ is the feature embedding dimensionality. Unlike MFD, MKD makes no Gaussian assumption, requires no covariance estimation, and is unbiased by construction, which makes it particularly suited to the small-sample regime ($n=30$ for feminization, $n=36$ for masculinization) considered here. 

For MFD, a dimensionality reduction step is required\footnote{Fitting a full covariance matrix requires the sample covariance to be full-rank, which demands $n > d = 512$. With only $n = 30$ or $n = 36$ generated samples, the sample covariance is rank-deficient, resulting in the matrix square root $(\mathbf{\Sigma}_g\mathbf{\Sigma}_t)^{1/2}$ being numerically ill-posed. Projecting onto the PCA components fitted on the real population reduces the covariance estimation problem to a well-conditioned system.}: we first apply principal component analysis (PCA) fitted on the full real population, retaining $32$ components that explain $99.8\%$ of variance. MKD is computed directly on the raw 512D embeddings without PCA, as it handles the $n \ll d$ regime by design. For both metrics, we report bootstrapped $95\%$ confidence intervals ($B=1{,}000$). As a reference, we report intra-class values obtained by randomly splitting each real population in half (averaged over $100$ splits), establishing a sampling noise floor. 
\begin{table}[htbp!]
    \centering
    \caption{Morphological Fr\'{e}chet Distance (MFD) and Morphological Kernel Distance (MKD, $\times 10^{3}$) between population pairs, computed from hold-out \texttt{SE-ResNet18} features. Lower values indicate greater distributional similarity. Intra-class values are averaged over $100$ random splits; bootstrapped $95\%$ CIs are shown in brackets.}
    \label{tab:distributional}
    \begin{tabular}{l c c}
        \toprule
        \textbf{Comparison} & \textbf{MFD} & \textbf{MKD} ($\times 10^{3}$) \\
        \midrule
        Real female vs.\ real female (intra-class) & $0.24 \pm 0.19$ & $0.15 \pm 1.89$ \\
        Real male vs.\ real male (intra-class) & $0.57 \pm 0.52$ & $0.62 \pm 9.30$ \\
        \midrule
        Real male vs.\ real female & $56.82$ & $662.38$ \\
        \midrule
        Gen.\ female (m$\rightarrow$f) vs.\ real female & $10.32\ [4.36,\, 18.11]$ & $\phantom{1}96.68\ [32.98,\, 117.93]$ \\
        Gen.\ male (f$\rightarrow$m) vs.\ real male & $10.99\ [6.57,\, 18.34]$ & $191.11\ [78.01,\, 375.47]$ \\
        \bottomrule
    \end{tabular}
\end{table}
\autoref{tab:distributional} reports both MFD and MKD metrics. The intra-class references establish the noise floor expected from finite sampling within a single distribution: MFD yields small positive values ($0.24$ and $0.57$), while MKD centers near zero ($0.15$ and $0.62$, $\times 10^{3}$). The inter-sex distances ($56.82$ and $662.38$, respectively) reflect the natural distributional separation in the learned feature space. After feminization (m~$\rightarrow$~f), MFD drops to $10.32$ ($\text{CI: } 4.36$--$18.11$) and MKD to $96.68$ ($\text{CI: } 32.98$--$117.93$), corresponding to reductions of $83\%$ and $86\%$ from the inter-sex baseline. The masculinization direction (f~$\rightarrow$~m) results in similar alignment under both metrics, with MFD at $10.99$ ($\text{CI: } 6.57$--$18.34$, a $82\%$ reduction), and MKD at $191.11$ ($\text{CI: } 78.01$--$375.47$, a $73\%$ reduction). The two metrics agree on the direction and magnitude of all comparisons. The fact that both estimators recover the same qualitative picture indicates that the observed distributional shift is robust to the choice of estimator and does not depend on assumptions about the shape of the feature distribution. Even the upper bounds of the bootstrap CIs for both metrics remain well below the inter-sex baseline, providing further evidence of the alignment of generated populations with their respective targets.

\subsubsection{Qualitative Assessment}
\begin{figure}[t]
    \centering
    \resizebox{\textwidth}{!}{
    \begin{tikzpicture}
            \node[] at (0, 0)     [anchor=south west]  {\includegraphics[height=4cm]{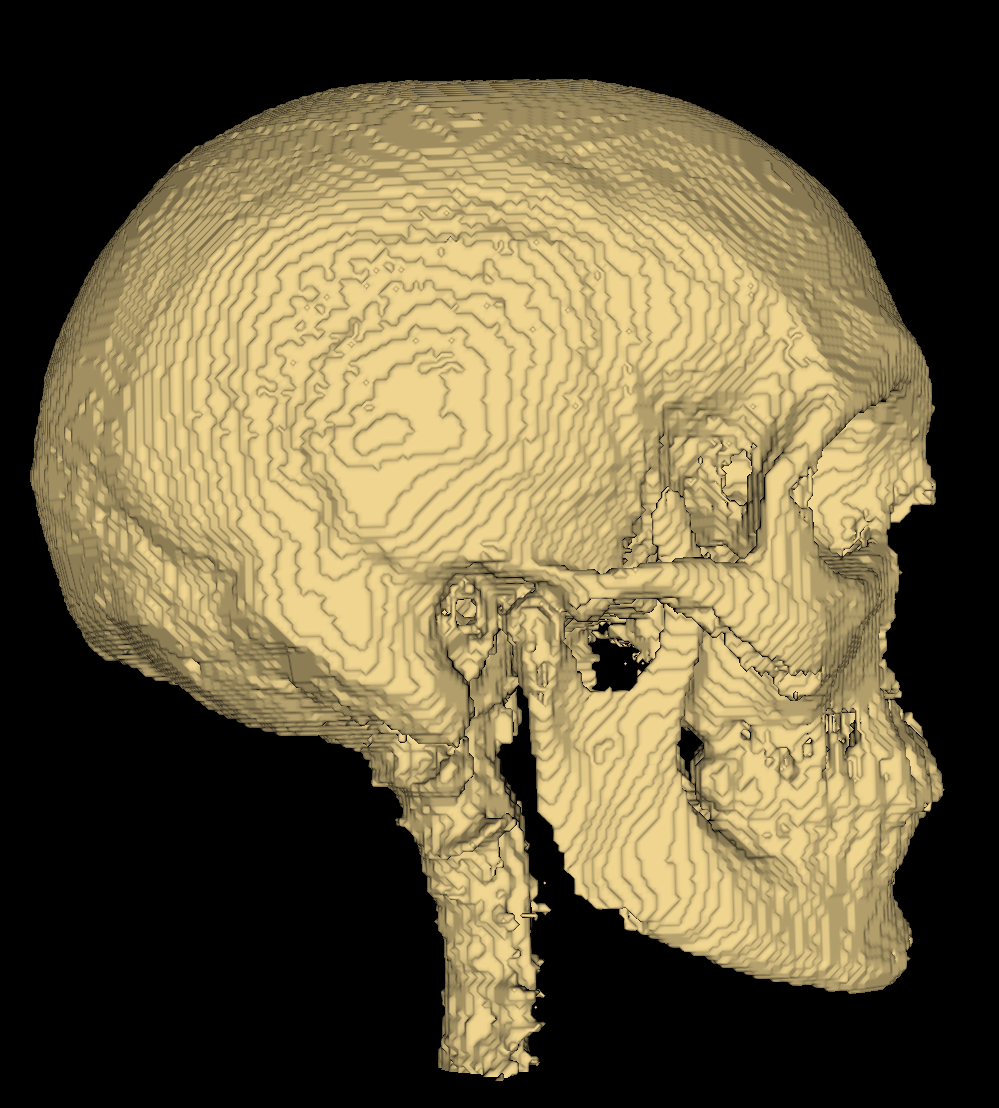}};
            \node[] at (1.8, 4.5) {\large Male (in)};
            \node[] at (3.6, 0)   [anchor=south west]  {\includegraphics[height=4cm]{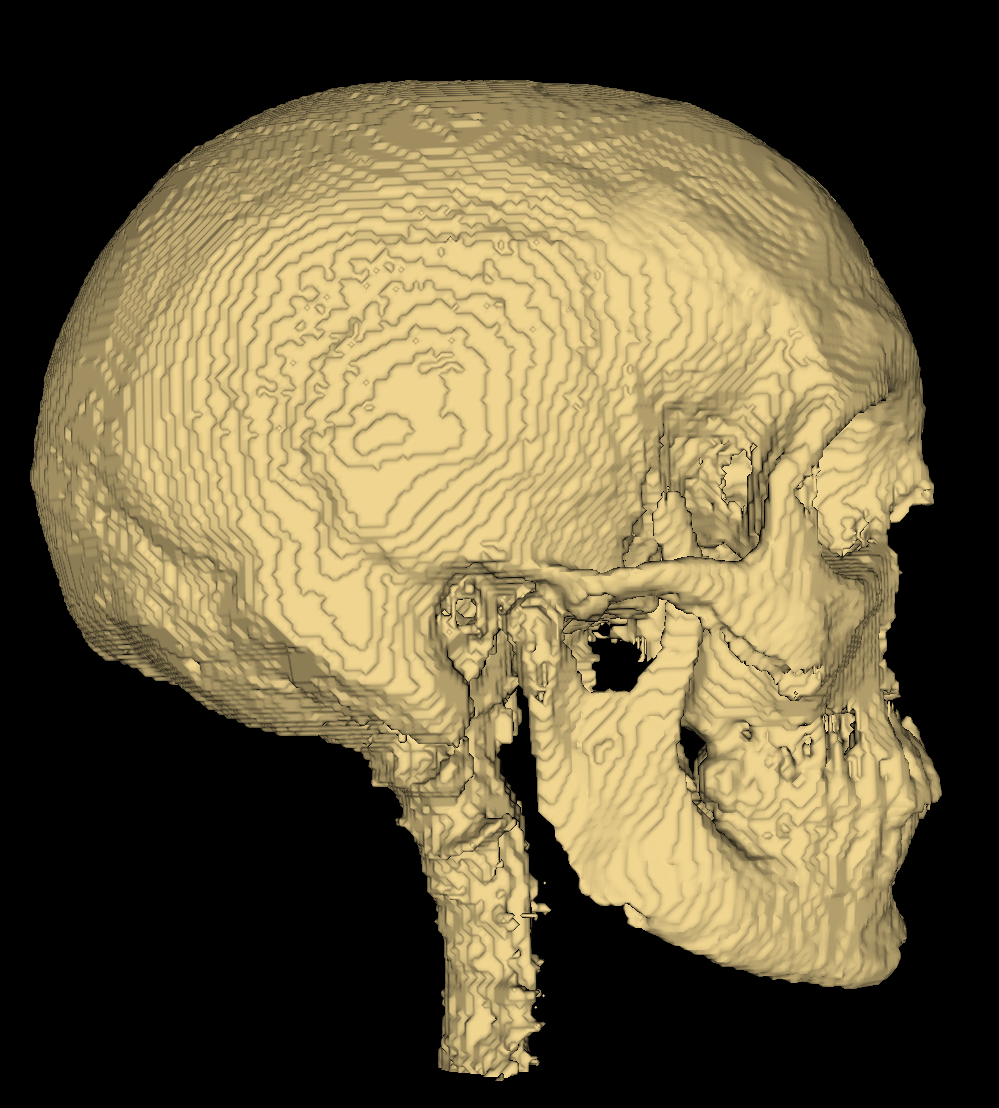}};
            \node[] at (5.4, 4.5) {\large Female (out)};
            \node[] at (7.2, 0)   [anchor=south west]  {\includegraphics[height=4cm]{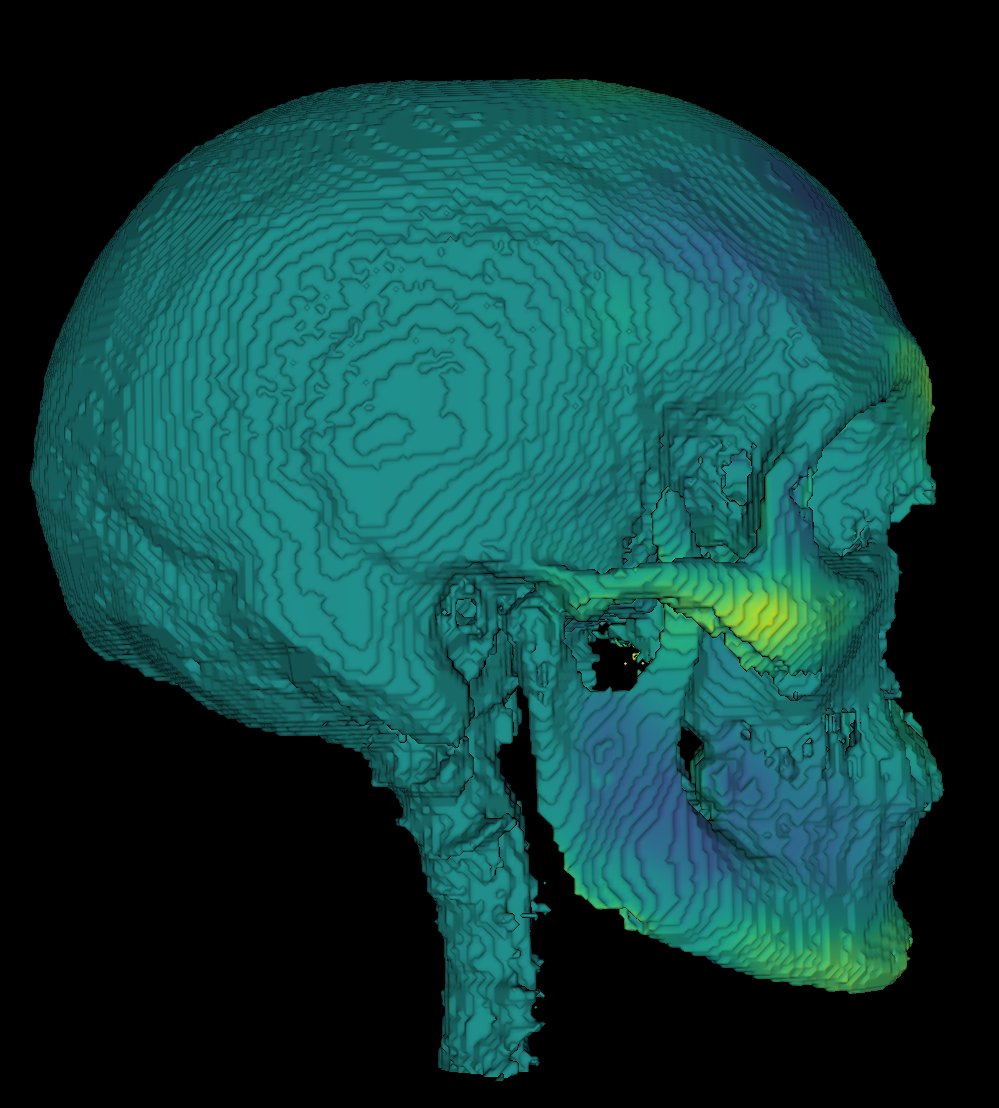}};
            \node[] at (9.0, 4.5) {\large Diff.};
            \draw[->] (2.9,4.5) -- (3.9,4.5);
            \node[rotate=90] at (-0.2, 2) {\large FFS};
            
            \node[] at (0, -5)     [anchor=south west]  {\includegraphics[height=4cm]{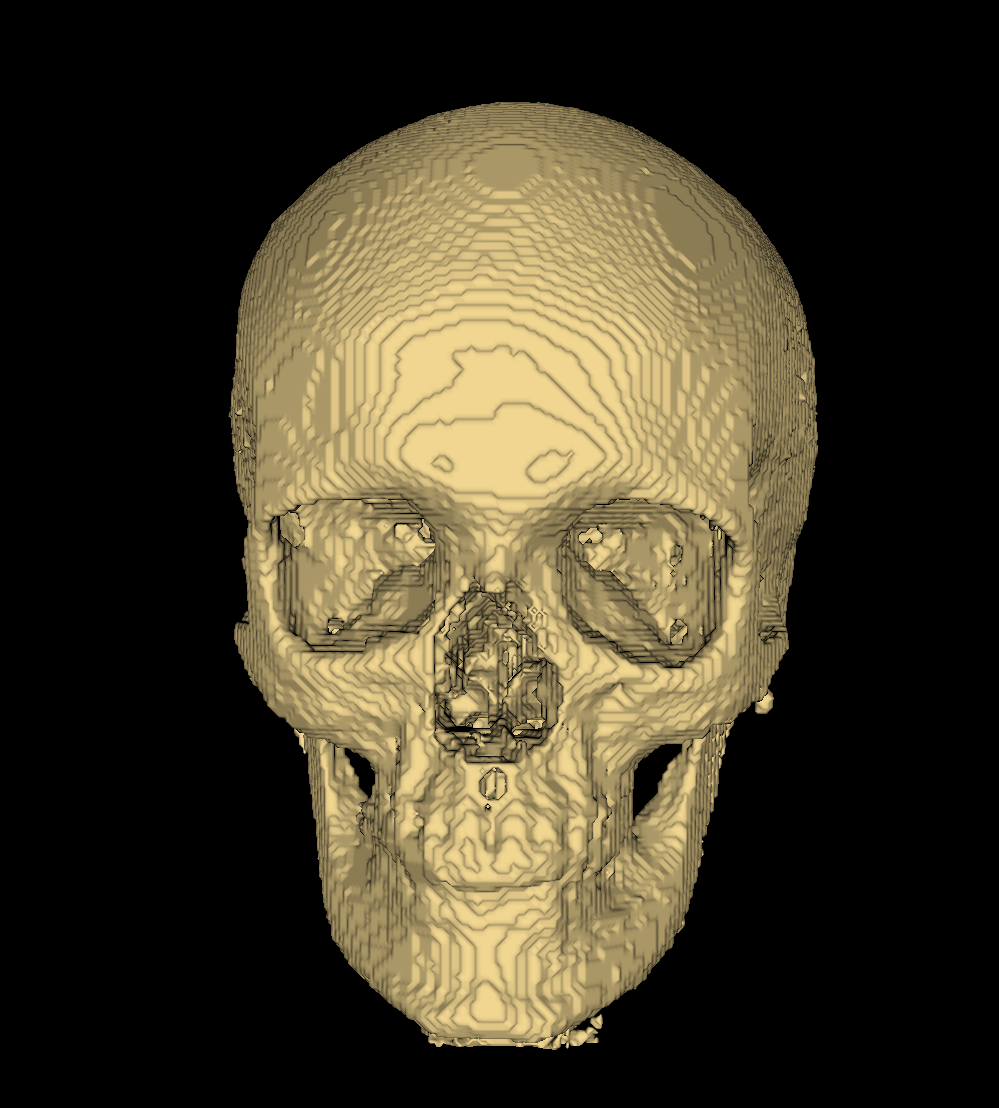}};
            \node[] at (1.8, -0.5) {\large Female (in)};
            \node[] at (3.6, -5)   [anchor=south west]  {\includegraphics[height=4cm]{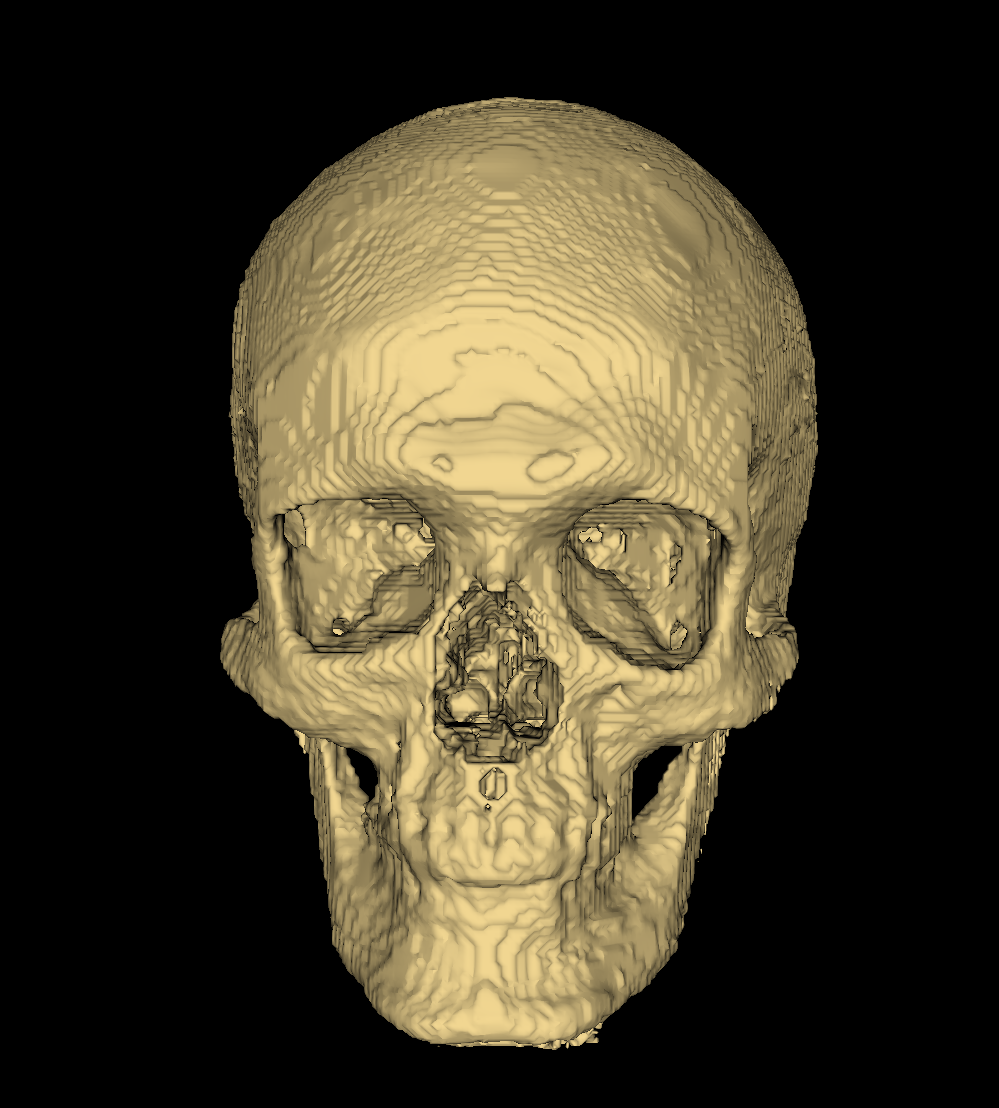}};
            \node[] at (5.4, -0.5) {\large Male (out)};
            \node[] at (7.2, -5)   [anchor=south west]  {\includegraphics[height=4cm]{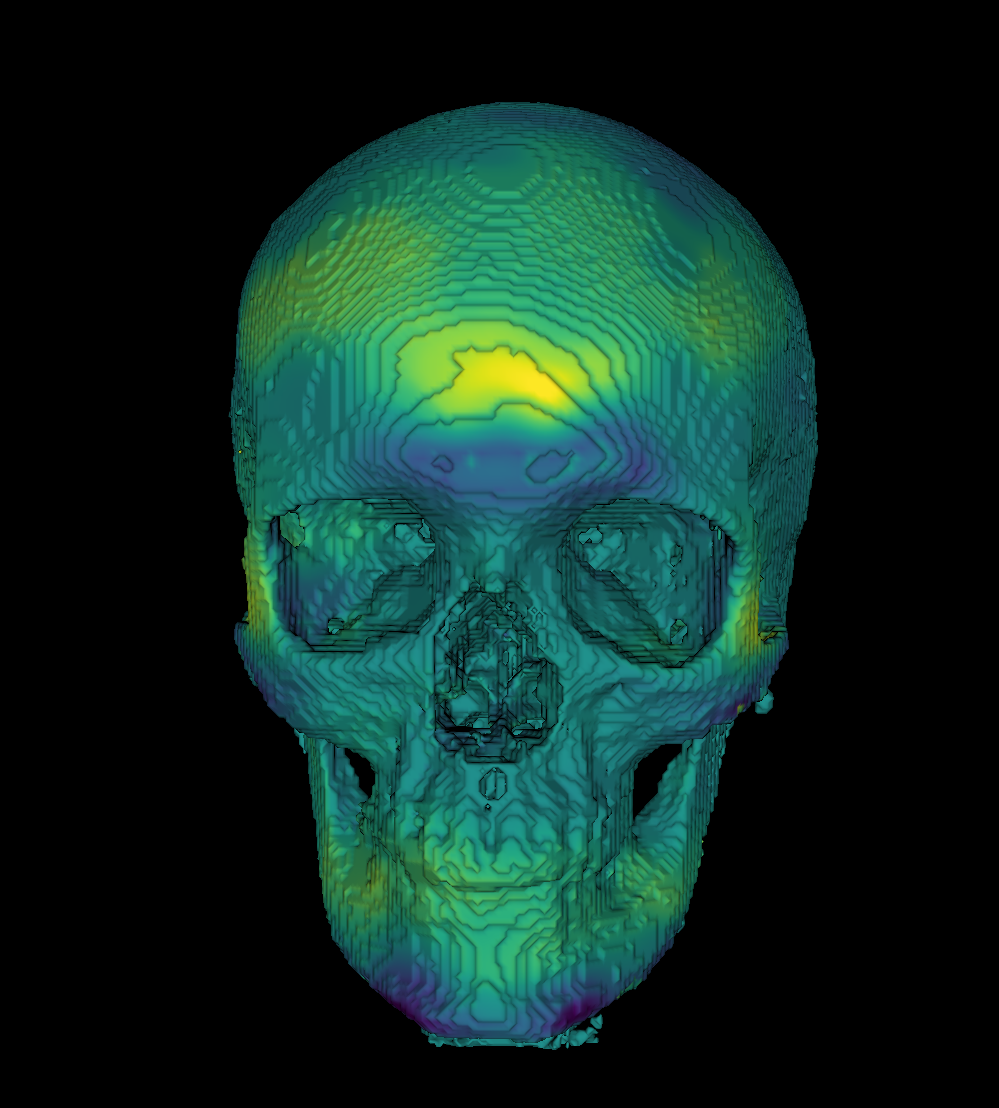}};
            \node[] at (9.0, -0.5)   {\large Diff.};
            \draw[->] (3.1,-0.5) -- (4.1,-0.5);
            \node[rotate=90] at (-0.2, -3) {\large FMS};
            \node[] at (11, -2.25) [anchor=south west]  {\includegraphics[height=4cm]{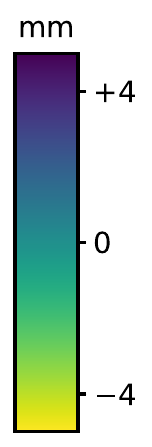}};
            
            \node[] at (13.5, -2.5)   [anchor=south west]  {\includegraphics[height=4cm]{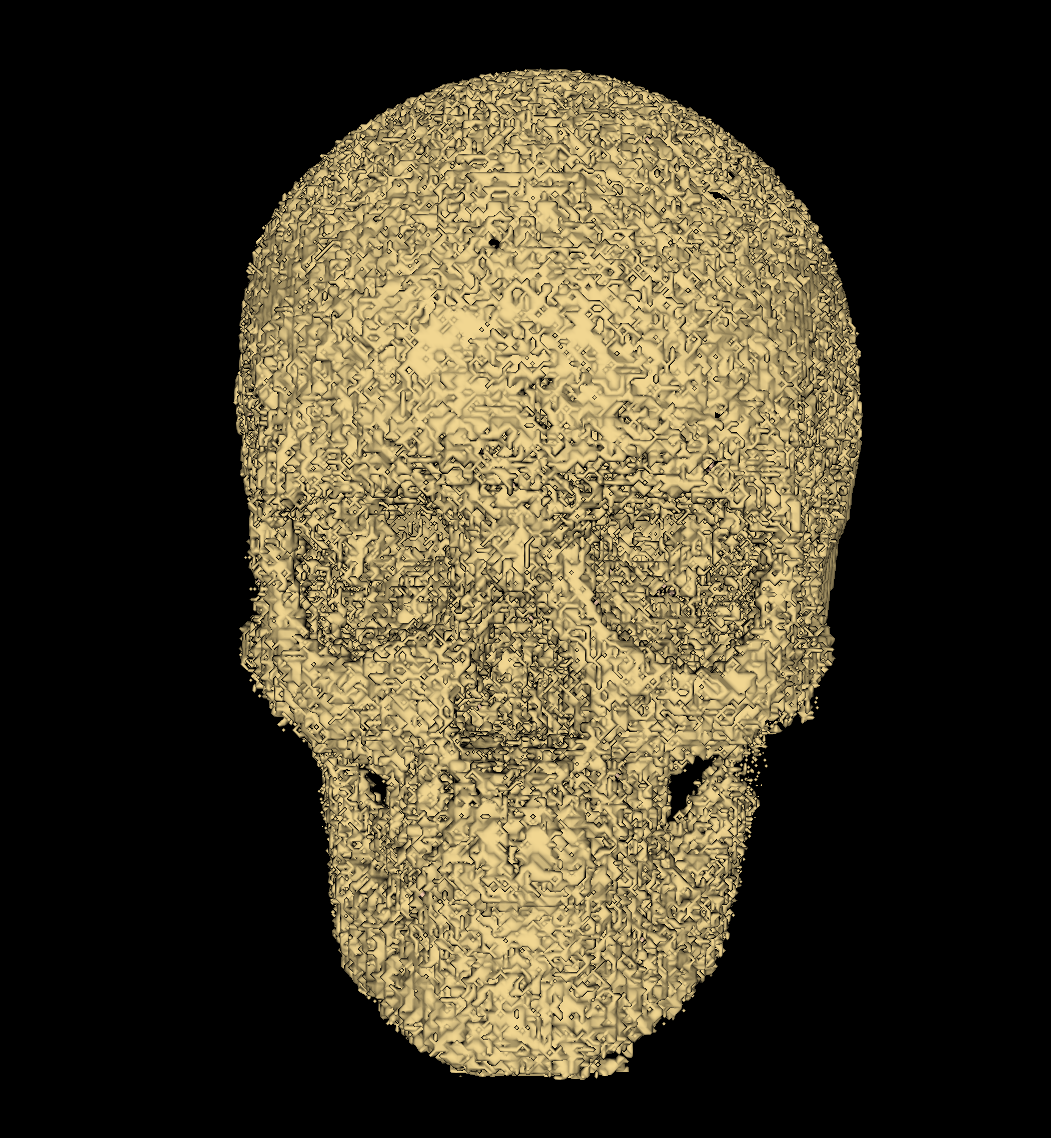}};
            \node[] at (15.5, 2)   {\large w/o B-spl. \& Reg.};
        \end{tikzpicture}}
    \caption{Example transformations. (\textit{Left, top}) FFS example from a lateral view. (\textit{Left, bottom}) FMS example from a frontal view. (\textit{Right}) Result without B-spline formulation and regularization.}
    \label{fig:example}
\end{figure}

The qualitative examples, shown in \autoref{fig:example}, demonstrate that the largest deformations concentrate in the chin, brow ridges, forehead, and the zygomatic bones: regions identified as sexually dimorphic in the anthropological literature~\cite{bannister2022sex}. Examining the two directions separately reveals a coherent and anatomically interpretable pattern. In the masculinization case~(FMS), the model produces a laterally expanded and generally thicker zygomatic bone, a wider chin, and a more pronounced brow ridge. The feminization case~(FFS) exhibits the inverse behaviour, with an attenuated brow ridge, a less anteriorly projecting chin, and a smaller, less laterally pronounced zygomatic bone. These per-direction changes are consistent with prior anthropological observations of cranial sexual dimorphism~\cite{del2023mapping}. This correspondence is notable given that no anatomical priors were imposed, meaning that the method recovers these regions purely from the learned classifier representations. This pattern is consistent across the dataset, not only the examples shown. \autoref{fig:example} (\textit{right}) further shows that removing the B-spline formulation and regularization terms causes the method to collapse to noise-like perturbation patterns, confirming the necessity of these constraints for producing anatomically coherent deformations.

\subsubsection{Human Perceptual Evaluation}
We further conduct a perceptual study with $N=11$ participants. Each rater was presented with $80$ skulls comprising $20$ real male, $20$ real female, $20$ generated male (f~$\rightarrow$~m), and $20$ generated female (m~$\rightarrow$~f) samples in randomized order. Raters were tasked with classifying the biological sex of each skull. Five skulls per category were shown two additional times to assess intra-rater agreement.
\begin{table}[b]
    \centering
    \caption{Perceptual study results ($N=11$). We report classification accuracy (mean $\pm$ SD across raters), inter-rater agreement ($\kappa_{\text{inter}}$, Fleiss') and intra-rater agreement ($\kappa_{\text{intra}}$, average pairwise Cohen's $\kappa$).}
    \label{tab:ratereval}
    \begin{tabular}{>{\raggedright\arraybackslash}p{2.2cm}|>{\centering\arraybackslash}p{2cm}>{\centering\arraybackslash}p{2cm}>{\centering\arraybackslash}p{2cm}}
        \toprule
        \textbf{Category} & \textbf{Acc.} & $\bm\kappa_{\textbf{inter}}$ & $\bm\kappa_{\textbf{intra}}$\\
        \midrule
        \text{Real} & $0.81 \pm 0.06$ & $0.51$ & $0.78$\\
        \text{Generated} & $0.37 \pm 0.08$ & $0.37$ & $0.43$\\
        \bottomrule
    \end{tabular}
\end{table}
As shown in \autoref{tab:ratereval}, raters correctly identified the sex of real skulls with $81\%$ accuracy ($\sim88\%$ is reported for expert anthropologists~\cite{walker2008sexing}) confirming that sexually dimorphic features are perceptible. For transformed skulls, accuracy dropped to $37\%$, well below chance, indicating that raters perceived the intended target sex in $63\%$ of cases. The $44\%$ drop in accuracy between real and generated skulls demonstrates that the pipeline produces a substantial and systematic perceptual shift. This effect was achieved despite the method only modifying anterior facial features while leaving posterior features and overall size unchanged. The lower agreement metrics for transformed skulls suggest that transformed morphologies are perceptually more ambiguous, consistent with the partial modification of sexually dimorphic features.
\section{Limitations and Future Work}
While the results presented in this paper are promising, there are limitations that should be considered when interpreting them. 

The dataset used for training and evaluation is relatively small and drawn from a multiple sclerosis (MS) patient cohort recruited in Switzerland. Consequently, it does not capture the full diversity of skull morphology across ethnicities, ages, and body types. As MS disproportionately affects women, the dataset exhibits a class imbalance skewed toward female subjects. Although we account for this imbalance during classifier training, it may still exert residual effects on the reported results.
Regarding image modality, MR-derived bone segmentations have been shown to agree closely with CT-derived models, and all segmentations were manually reviewed. Nevertheless, CT imaging remains the reference modality in FFS planning. Assembling a large-scale, multi-center CT dataset therefore is the primary objective of our future work.

We additionally emphasize that this work only addresses a subset of the full FFS planning pipeline. The presented results focus on bone morphology modifications, which, consistent with standard surgical practice, precede soft tissue changes. Extending the pipeline to include soft tissue simulation on the modified bony structures, as well as the derivation of patient-specific surgical guides (e.g., cutting guides) from the generated results, are important directions that remain open for future work.

\section{Conclusion}
We present \textbf{AutoFFS}, a data-driven framework for facial feminization surgery planning that reframes \emph{targeted adversarial attacks} as a mechanism for clinically meaningful shape editing. By optimizing a regularized B-spline deformation field against an ensemble of pre-trained sex classifiers, the method repurposes adversarial optimization as a tool for generating personalized, anatomically plausible counterfactual skull morphologies. The classifier-based evaluation, population-level distributional metrics, and a human perceptual study suggest that data-driven planning for FFS is a viable research direction with meaningful clinical potential, indicated by both qualitative and quantitative results. We hope this work not only serves as a technical contribution, but also as an invitation to the community to engage with this largely overlooked but clinically and socially relevant problem.

\begin{credits}
\subsubsection{\ackname} We thank Cristina Granziera and Lester Melie-Garcia for providing access to the dataset and Martin Styner for helpful discussions around the evaluation of the method. We would further like to thank Judith Zecha, Ruud Schreurs, Juliana Sabelis and Hugo Beltman for inspiration and helpful discussions throughout the project. The Werner Siemens Foundation financially supported this work through the MIRACLE II project.

\subsubsection{\discintname}
The authors have no competing interests to declare that are relevant to the content of this article.
\end{credits}

\bibliographystyle{splncs04.bst}
\bibliography{bibliography.bib}
\end{document}